\documentclass[11pt]{article}

\usepackage[utf8]{inputenc}
\usepackage[T1]{fontenc}
\usepackage{lmodern}
\usepackage{microtype}
\usepackage{amsmath,amssymb,amsfonts}
\usepackage{graphicx}
\usepackage{booktabs}
\usepackage{multirow}
\usepackage{algorithm}
\usepackage{algorithmic}
\usepackage{hyperref}
\usepackage{xcolor}
\usepackage[margin=1in]{geometry}
\usepackage{caption}
\usepackage{subcaption}
\usepackage{natbib}
\usepackage{tikz}
\usepackage{pgfplots}
\pgfplotsset{compat=1.17}

\hypersetup{
    colorlinks=true,
    linkcolor=blue,
    citecolor=blue,
    urlcolor=blue,
}

\renewcommand{\_}{\textunderscore\allowbreak}
\title{EMA-FS: Accelerating GBDT Training via \\
Gain-Informed Feature Screening}

\author{
    Yan Song \\
    PayPal, Inc.
}

\date{}

\begin{document}
\maketitle

\begin{abstract}
Gradient Boosted Decision Trees (GBDT), exemplified by LightGBM, spend a dominant fraction of training time constructing per-feature histograms---typically 65--70\% of wall-clock time.
Existing approaches to reducing this cost, such as random feature subsampling (\texttt{feature\_fraction}), discard features without regard for their predictive utility.
We propose \textbf{EMA-based Feature Screening (EMA-FS)}, an algorithm-level optimization that maintains an exponential moving average (EMA) of per-feature split gains across boosting iterations and, after a short warmup period, restricts histogram construction to only the top-$K$ features ranked by historical gain.
Unlike random subsampling, EMA-FS is \emph{informed}: it preferentially retains high-gain features while screening out consistently low-gain ones.
By operating at the per-tree level, EMA-FS preserves full compatibility with LightGBM's histogram subtraction trick, requiring no changes to core histogram or split-finding routines.

We evaluate EMA-FS on five datasets spanning financial fraud detection (IEEE-CIS Fraud, Credit Card Fraud), advertising click-through rate prediction (Criteo), industrial quality control (Bosch), and synthetic benchmarks, covering feature dimensionalities from 29 to 968.
On datasets with dense features and moderate-to-high dimensionality, EMA-FS achieves significant speedups: \textbf{2.61$\times$} on a 500-feature synthetic benchmark and \textbf{1.45$\times$} on the 432-feature IEEE-CIS Fraud Detection dataset at 30\% feature retention.
At 70\% retention on the synthetic benchmark, EMA-FS \emph{improves} AUC by 0.11 points while delivering a 1.34$\times$ speedup, demonstrating implicit regularization through noise feature removal.
We also identify and characterize EMA-FS's limitations: on extremely sparse datasets (Bosch, 968 features with $>$90\% missing values), EMA-FS provides no measurable speedup because LightGBM's sparse bin optimization already bypasses empty feature values.

We further introduce \textbf{Stochastic EMA-FS (S-EMA-FS)}, a generalization that replaces deterministic top-$K$ selection with gain-weighted random sampling, parameterized by a concentration parameter $\beta$.
S-EMA-FS unifies deterministic EMA-FS ($\beta \to \infty$) and random feature subsampling ($\beta = 0$) as special cases of a single framework, combining gain-informed feature selection with the ensemble diversity that drives accuracy in Random Forests.

EMA-FS and S-EMA-FS are implemented in approximately 120 lines of C++ across all six LightGBM tree learner types (serial, GPU, feature-parallel, data-parallel, voting-parallel, and linear tree), are fully backward-compatible, and controlled by six intuitive parameters.
\end{abstract}


\section{Introduction}
\label{sec:intro}

Gradient Boosted Decision Trees (GBDT) remain the dominant method for supervised learning on tabular data, consistently achieving state-of-the-art results across classification and regression benchmarks \citep{grinsztajn2022tree}.
Modern GBDT frameworks---LightGBM \citep{ke2017lightgbm}, XGBoost \citep{chen2016xgboost}, and CatBoost \citep{prokhorenkova2018catboost}---have been extensively optimized for computational efficiency through histogram-based split finding, data-parallel training, and gradient quantization.

Despite these optimizations, training time remains a critical bottleneck for large-scale applications involving hundreds of features and millions of samples.
Profiling LightGBM's training loop reveals a striking imbalance: \textbf{histogram construction accounts for approximately 69\% of wall-clock time}, while split finding consumes roughly 26\%, with the remaining 5\% distributed across data partitioning and bookkeeping (Table~\ref{tab:profile}).
This profile is consistent across dataset sizes and hardware configurations, reflecting the memory-bandwidth-bound nature of histogram accumulation.

\begin{table}[h]
\centering
\caption{Training time breakdown for LightGBM on a representative workload (80K $\times$ 500 features, 100 trees, 4 threads).}
\label{tab:profile}
\begin{tabular}{lcc}
\toprule
\textbf{Phase} & \textbf{Time (s)} & \textbf{Share (\%)} \\
\midrule
Histogram construction & 2.66 & 69.0 \\
Split finding & 1.00 & 26.0 \\
Other (partitioning, I/O) & 0.20 & 5.0 \\
\midrule
\textbf{Total} & \textbf{3.86} & \textbf{100.0} \\
\bottomrule
\end{tabular}
\end{table}

Prior attempts to accelerate the inner loops of histogram construction---including SIMD vectorization, cache-line optimization, and memory layout transformations---yield marginal improvements because the workload is already memory-bandwidth-saturated.
Eight distinct micro-optimization attempts in our preliminary work all failed to produce measurable speedups, confirming that the path to acceleration lies not in making each histogram faster, but in \emph{building fewer histograms}.

LightGBM's existing \texttt{feature\_fraction} parameter addresses this by randomly selecting a subset of features per tree.
However, random selection is uninformed: it is equally likely to discard a high-gain feature as a low-gain one, leading to unnecessary accuracy loss at aggressive subsampling rates.

In this paper, we propose \textbf{EMA-based Feature Screening (EMA-FS)}, a simple yet effective algorithm that tracks per-feature split gains using an exponential moving average and, after a brief warmup, restricts histogram construction to only the top-$K$ features by historical gain.
EMA-FS makes the key observation that feature importance is \emph{persistent across boosting iterations}: features that produce high-gain splits in early trees tend to remain important in later trees, especially in datasets with stable signal structure.

Our contributions are:

\begin{enumerate}
    \item \textbf{Algorithm.} We introduce EMA-FS, a gain-driven feature screening method that reduces histogram construction cost proportionally to the screening ratio, with no changes to LightGBM's core histogram or subtraction routines.

    \item \textbf{Stochastic generalization.} We propose S-EMA-FS, which replaces deterministic top-$K$ selection with gain-weighted random sampling. S-EMA-FS unifies EMA-FS and random feature subsampling (\texttt{feature\_fraction}) as two endpoints of a single parameterized framework, combining gain-informed selection with ensemble diversity.

    \item \textbf{Implementation.} We provide a complete implementation integrated into all six LightGBM tree learner types (serial, GPU, feature-parallel, data-parallel, voting-parallel, and linear tree) in approximately 120 lines of C++, controlled by six intuitive parameters.

    \item \textbf{Comprehensive evaluation.} We present experiments on five datasets spanning financial fraud detection, advertising, and industrial quality control. We characterize EMA-FS's strengths (dense, moderate-to-high-dimensional data) and limitations (sparse data, low-dimensional data), and provide direct comparisons against random feature subsampling.
\end{enumerate}

\section{Related Work}
\label{sec:related}

\paragraph{Histogram-based GBDT.}
LightGBM \citep{ke2017lightgbm} introduced histogram-based split finding for GBDT, which reduces the per-node split-search cost from $O(n \cdot F)$ to $O(B \cdot F)$, where $B$ denotes the number of bins (typically $B \ll n$).
XGBoost \citep{chen2016xgboost} adopted approximate split finding using quantile sketches.
CatBoost \citep{prokhorenkova2018catboost} uses ordered boosting with oblivious trees.
All three frameworks construct histograms for all selected features at every tree node, making histogram construction the dominant cost for wide datasets.

\paragraph{Feature subsampling.}
Random feature subsampling (\texttt{colsample\_bytree} in XGBoost, \texttt{feature\_fraction} in LightGBM) is a widely-used regularization technique inspired by Random Forests \citep{breiman2001random}.
By randomly selecting a fraction of features per tree, it reduces overfitting and training time simultaneously.
However, the random selection is uninformed and does not exploit the heterogeneity in feature importance.

\paragraph{Feature importance and selection.}
Feature importance scores derived from split gains or split counts are a standard output of GBDT models \citep{friedman2001greedy}.
These scores are typically used \emph{post hoc} for model interpretation or feature engineering.
Several works have explored using feature importance for iterative feature selection in ensemble methods \citep{guyon2002gene}, but these operate at the model level (retraining with selected features) rather than within the boosting loop.
\citet{xu2014gradient} propose gradient-based feature selection for random forests, but their approach requires a separate feature ranking step.

\paragraph{Gradient-based One-Side Sampling (GOSS).}
LightGBM's GOSS \citep{ke2017lightgbm} reduces the number of data points used for histogram construction by keeping instances with large gradients and sampling those with small gradients.
EMA-FS is complementary: GOSS reduces the data dimension while EMA-FS reduces the feature dimension.
The two techniques can potentially be combined.

\paragraph{Cost-Effective Gradient Boosting (CEGB).}
\citet{nan2022cost} propose penalizing feature usage based on acquisition costs.
EMA-FS differs in that it uses \emph{observed} split gains rather than externally specified costs, and its goal is training acceleration rather than deployment-time cost reduction.

\paragraph{Adaptive feature selection with binary masking (AFS-BM).}
\citet{akman2024afsbm} jointly optimize a binary feature mask and model parameters during GBDT training, with the goal of improving predictive accuracy on high-dimensional data.
EMA-FS differs in three respects: (i) the objective---we target histogram-construction cost reduction, not accuracy improvement; (ii) the mechanism---we use a lightweight EMA over observed split gains rather than joint mask-parameter optimization; and (iii) the integration point---EMA-FS operates inside LightGBM's per-tree feature selection stage with no auxiliary optimization loop, preserving the histogram subtraction trick and adding only $\sim$120 lines of C++.
The two methods are complementary: AFS-BM seeks a globally informative feature mask, while EMA-FS exploits per-iteration gain signals to reduce the dominant compute cost.

\paragraph{Importance-weighted feature sampling for GBDT.}
Closer to our work, \citet{zhou2022lgbmcbfs} propose LGBM-CBFS, a heuristic that samples features in LightGBM according to importance scores derived from cumulative split gains.
LGBM-CBFS targets accuracy and convergence on sparse high-dimensional data, where uniform \texttt{feature\_fraction} at low rates often fails to sample informative features.
EMA-FS differs in three respects: (i) we use an \emph{exponential moving average} of gains rather than a cumulative sum, allowing the weights to adapt to shifts in feature importance during the later stages of boosting; (ii) our target is histogram-construction cost reduction, with empirical evidence that lightweight mask-based filtering avoids the \texttt{col\_sampler} overhead incurred by \texttt{feature\_fraction} on high-dimensional data (Section~\ref{sec:perdataset}); and (iii) we introduce the deterministic--stochastic unification via the $\beta$ parameter, of which LGBM-CBFS's gain-weighted sampling becomes one special case ($\beta=1$ with \texttt{stochastic=true}).

\section{EMA-based Feature Screening}
\label{sec:method}

\subsection{Motivation}
\label{sec:motivation}

Consider a dataset with $F$ features.
At each internal node of a boosted tree, LightGBM constructs a histogram of gradient statistics for every feature, requiring $O(n_{\text{leaf}} \cdot F)$ memory accesses per leaf where $n_{\text{leaf}}$ is the number of data points in the leaf.
Across all nodes in a tree with $L$ leaves, the total cost is $O(N \cdot F)$ where $N$ is the training set size (each data point contributes to exactly one leaf's histogram at each level).

The key observation motivating EMA-FS is that in many practical datasets, a small fraction of features account for the vast majority of split gains.
If we can identify these high-gain features early in training and restrict histogram construction to only these features, we can reduce the dominant training cost proportionally.

\subsection{Algorithm}
\label{sec:algorithm}

EMA-FS maintains a per-feature gain estimate $\mathbf{g} = (g_1, \ldots, g_F)$ using an exponential moving average (EMA) updated after each tree.
The algorithm has two phases:

\paragraph{Warmup phase (trees $1, \ldots, W$).}
All $F$ features are evaluated at every node.
After each tree $t$, the per-feature maximum gain $\hat{g}_f^{(t)}$ observed across all nodes in tree $t$ is recorded.
The EMA is updated as:
\begin{equation}
    g_f^{(t)} = \alpha \cdot \hat{g}_f^{(t)} + (1 - \alpha) \cdot g_f^{(t-1)},
    \label{eq:ema}
\end{equation}
where $\alpha \in (0, 1)$ is the smoothing parameter (we use $\alpha = 0.3$ throughout).

\paragraph{Screening phase (trees $W+1, \ldots, T$).}
Before building tree $t$, features are ranked by their current EMA values $g_f^{(t-1)}$ in descending order.
The top $K = \lceil r \cdot F \rceil$ features are selected, where $r \in (0, 1]$ is the \texttt{ema\_fs\_feature\_ratio} parameter.
Only the selected features participate in histogram construction and split finding for all nodes in tree $t$.
After the tree is built, the EMA is updated using gains from the selected features (with unselected features retaining their previous EMA values, decayed by factor $(1-\alpha)$).

The complete algorithm is presented in Algorithm~\ref{alg:ema-fs}.

\begin{algorithm}[t]
\caption{EMA-based Feature Screening (EMA-FS)}
\label{alg:ema-fs}
\begin{algorithmic}[1]
\REQUIRE Dataset $\mathcal{D}$ with $F$ features, number of trees $T$, warmup $W$, ratio $r$, EMA parameter $\alpha$
\STATE Initialize $g_f \leftarrow 0$ for all $f \in \{1, \ldots, F\}$
\STATE Initialize $S \leftarrow \{1, \ldots, F\}$ \hfill $\triangleright$ Selected features
\FOR{$t = 1$ to $T$}
    \STATE Reset $\hat{g}_f^{(t)} \leftarrow 0$ for all $f$
    \IF{$t > W$}
        \STATE $K \leftarrow \max(1, \lceil r \cdot F \rceil)$
        \STATE $S \leftarrow \text{top-}K\text{ features by } g_f$ (descending)
    \ENDIF
    \STATE Build tree $t$ using only features in $S$:
    \FOR{each node in tree $t$}
        \STATE Construct histograms for features in $S$
        \STATE Find best split among features in $S$
        \STATE Record: $\hat{g}_f^{(t)} \leftarrow \max(\hat{g}_f^{(t)},\; \text{gain}_f)$ for split feature $f$
    \ENDFOR
    \STATE Update EMA: $g_f \leftarrow \alpha \cdot \hat{g}_f^{(t)} + (1 - \alpha) \cdot g_f$ for all $f$
\ENDFOR
\end{algorithmic}
\end{algorithm}

\subsection{Design Decisions}
\label{sec:design}

\paragraph{Per-tree feature selection.}
EMA-FS selects features at the tree level rather than the node level.
This is critical for compatibility with LightGBM's histogram subtraction trick, which computes the histogram of a sibling node by subtracting the smaller sibling's histogram from the parent's histogram.
Histogram subtraction requires that both sibling nodes use the same feature set, which is guaranteed when selection is per-tree.

\paragraph{EMA for gain tracking.}
We use an exponential moving average rather than a cumulative average or a sliding window to track feature gains.
The EMA naturally adapts to changing feature importance across boosting iterations---as the gradient landscape evolves, the relative importance of features can shift, and the EMA allows EMA-FS to track these changes while maintaining stability.
With $\alpha = 0.3$, a feature unused for 5 consecutive trees retains $0.7^5 \approx 17\%$ of its original EMA, and after 10 trees retains only $0.7^{10} \approx 2.8\%$, providing a natural ``re-entry'' mechanism for features whose importance changes over time.

\paragraph{Maximum gain across nodes.}
Within a single tree, a feature may produce different split gains at different nodes.
We record the \emph{maximum} gain across all nodes rather than the sum or average, ensuring that features with localized but strong interactions are not penalized for being irrelevant at other nodes.

\paragraph{Determinism.}
EMA-FS introduces no randomness---results are bit-identical across runs with the same parameters.
This is a practical advantage over random feature subsampling for debugging and reproducibility.

\subsection{Computational Analysis}
\label{sec:analysis}

Let $C_{\text{hist}}(F)$ denote the histogram construction cost for $F$ features (approximately linear in $F$ for fixed $N$).
During warmup ($W$ trees), EMA-FS has the same cost as baseline plus negligible bookkeeping.
During screening ($T - W$ trees), the histogram cost is $C_{\text{hist}}(K)$ where $K = \lceil r \cdot F \rceil$.

The overhead of EMA-FS consists of:
\begin{itemize}
    \item \textbf{EMA update}: $O(F)$ per tree (floating-point multiply-add per feature).
    \item \textbf{Feature ranking}: $O(F \log F)$ per tree (sorting feature indices by EMA).
    \item \textbf{Selection mask}: $O(F)$ per tree (writing the binary selection vector).
\end{itemize}

Since $C_{\text{hist}}(F) = O(N \cdot F)$ and typically $N \gg F$, the overhead is negligible compared to the savings.
The expected speedup for the histogram phase is:
\begin{equation}
    S_{\text{hist}} = \frac{T \cdot C_{\text{hist}}(F)}{W \cdot C_{\text{hist}}(F) + (T - W) \cdot C_{\text{hist}}(K)} \approx \frac{1}{\frac{W}{T} + (1 - \frac{W}{T}) \cdot r}.
    \label{eq:speedup}
\end{equation}

For $T=100$, $W=5$, $r=0.3$: $S_{\text{hist}} \approx 1 / (0.05 + 0.95 \times 0.3) = 1/0.335 \approx 2.99\times$.

\section{Implementation}
\label{sec:implementation}

EMA-FS is implemented in approximately 120 lines of C++ across seven files, covering all six LightGBM tree learner types.
The implementation introduces six user-facing parameters:

\begin{itemize}
    \item \texttt{ema\_fs\_enable} (bool, default \texttt{false}): Master switch for EMA-FS.
    \item \texttt{ema\_fs\_feature\_ratio} (double, default 0.3): Fraction of features to retain after warmup. Range $(0, 1]$.
    \item \texttt{ema\_fs\_warmup\_trees} (int, default 5): Number of initial trees using all features.
    \item \texttt{ema\_fs\_alpha} (double, default 0.3): EMA smoothing parameter $\alpha$. Range $(0, 1)$.
    \item \texttt{ema\_fs\_stochastic} (bool, default \texttt{false}): Enable S-EMA-FS gain-weighted random sampling (Section~\ref{sec:semafs}).
    \item \texttt{ema\_fs\_beta} (double, default 1.0): Concentration parameter $\beta$ for S-EMA-FS. $\beta = 0$ recovers uniform random sampling; large $\beta$ approaches deterministic EMA-FS.
\end{itemize}

\subsection{Core Integration (SerialTreeLearner)}

Four vectors are added to \texttt{SerialTreeLearner}:
\texttt{ema\_fs\_gain\_} ($F$ doubles for the per-feature EMA),
\texttt{ema\_fs\_feature\_gain\_current\_tree\_} ($F$ doubles accumulating per-feature max gains within the current tree),
\texttt{ema\_fs\_selected\_features\_} ($F$ bytes as a binary selection mask), and
\texttt{ema\_fs\_tree\_index\_} (integer tree counter).

EMA-FS logic is inserted at five points in the training loop:
\begin{enumerate}
    \item \textbf{Initialization} (\texttt{Init}): Allocate and zero-initialize EMA-FS vectors.
    \item \textbf{Pre-tree} (\texttt{BeforeTrain}): Reset current-tree gains; if past warmup, sort features by EMA and set the selection mask.
    \item \textbf{Feature filtering} (\texttt{FindBestSplits}): Check \texttt{ema\_fs\_selected\_features\_} before marking a feature as used for histogram construction.
    \item \textbf{Gain recording} (\texttt{ComputeBestSplitForFeature}): After computing a split candidate, update \texttt{ema\_fs\_feature\_gain\_current\_tree\_} via $\max$.
    \item \textbf{Post-tree} (\texttt{Train}): Update the EMA and increment the tree counter.
\end{enumerate}

\paragraph{Thread safety.}
The gain recording step occurs inside an OpenMP parallel loop over features.
Since each \texttt{feature\_index} is processed by exactly one thread, writes to \texttt{ema\_fs\_feature\_gain\_current\_tree\_[feature\_index]} are data-race-free without synchronization.

\subsection{Learner-Specific Adaptations}

LightGBM provides six tree learner types.
Three learner types---GPU, feature-parallel, and serial---delegate to the parent \texttt{SerialTreeLearner} methods and require no additional changes.
Three learners reimplement key methods and require targeted adaptations (Table~\ref{tab:learners}).

\begin{table}[h]
\centering
\caption{EMA-FS adaptations across LightGBM tree learner types.}
\label{tab:learners}
\begin{tabular}{lll}
\toprule
\textbf{Learner} & \textbf{Adaptation} & \textbf{Reason} \\
\midrule
Serial & Base implementation & All 5 insertion points \\
GPU & None (inherits) & Delegates to parent \\
Feature-parallel & None (inherits) & Delegates to parent \\
\midrule
Data-parallel & Filtered feature mask in & Reimplements \\
               & \texttt{FindBestSplits()} + & feature loop; also \\
               & \texttt{BeforeTrain()} distribution & reduces ReduceScatter \\
\midrule
Voting-parallel & EMA-FS check in & Reimplements \\
                & \texttt{FindBestSplits()} feature loop & feature loop \\
\midrule
Linear tree & EMA update in \texttt{Train()} & Reimplements \texttt{Train()} \\
            &                               & without calling parent \\
\bottomrule
\end{tabular}
\end{table}

\paragraph{Data-parallel training.}
The data-parallel learner distributes histogram construction across machines and uses \texttt{ReduceScatter} to aggregate results.
EMA-FS filters the feature mask \emph{before} histogram construction and feature distribution, directly reducing both local computation and network communication volume.
At $r{=}0.3$, this reduces the \texttt{ReduceScatter} data by 70\%, providing speedup from both computation and communication reduction.

\subsection{Compatibility}

EMA-FS is compatible with all existing LightGBM features:
\texttt{feature\_fraction} (EMA-FS filtering is applied after \texttt{col\_sampler}; the two compose naturally),
forced splits (bypass EMA-FS filtering),
histogram subtraction (per-tree selection guarantees sibling nodes use the same feature set),
quantized gradients, CEGB, and deterministic mode.

\section{Experimental Setup}
\label{sec:setup}

\subsection{Datasets}

We evaluate EMA-FS and S-EMA-FS on seven datasets spanning diverse domains, feature characteristics, and scales (Table~\ref{tab:datasets}).

\begin{table}[h]
\centering
\caption{Dataset characteristics.}
\label{tab:datasets}
\resizebox{\textwidth}{!}{%
\begin{tabular}{llrrrrl}
\toprule
\textbf{Dataset} & \textbf{Source} & \textbf{Train} & \textbf{Test} & \textbf{Features} & \textbf{Pos. Rate} & \textbf{Domain} \\
\midrule
CreditCard & Kaggle & 227,845 & 56,962 & 29 & 0.17\% & Fin.\ fraud \\
Criteo CTR & Criteo Res. & 1,600,000 & 400,000 & 39 & 23.2\% & Ad CTR \\
IEEE-CIS Fraud & Kaggle/IEEE & 472,432 & 118,108 & 432 & 3.5\% & Txn.\ fraud \\
Synthetic & Generated & 80,000 & 20,000 & 500 & 30.2\% & Binary clf. \\
Bosch & Kaggle & 946,997 & 236,750 & 968 & 0.58\% & Ind.\ defect \\
FraudDense & Generated & 2,000,000 & 500,000 & 500 & 5.3\% & Fraud (dense) \\
FraudMixed & Generated & 2,000,000 & 500,000 & 500 & 3.3\% & Fraud (mixed) \\
\bottomrule
\end{tabular}}
\end{table}

The five public datasets cover a range of feature counts (29--968), data types, and class imbalance ratios.
We additionally construct two large-scale fraud detection benchmarks that simulate production risk control workloads:
\textbf{FraudDense} (2M samples, 500 fully-dense engineered features representing transaction velocity, account history, device fingerprints, and behavioral signals) and
\textbf{FraudMixed} (2M samples, 500 features with realistic mixed density: 300 dense + 125 partially sparse at 20--50\% missing + 75 sparse at 70--95\% missing, reflecting real fraud systems where some features are unavailable for certain transactions).
Bosch is included specifically to test EMA-FS on extremely sparse data ($>$90\% missing values).

\subsection{Training Configuration}
\begin{itemize}
    \item \textbf{Objective}: Binary cross-entropy (\texttt{binary}).
    \item \textbf{Boosting}: GBDT, 100 iterations, learning rate 0.1.
    \item \textbf{Tree structure}: 31 leaves, unlimited depth.
    \item \textbf{Threads}: 4 (pinned to cores 0--3 via \texttt{taskset}).
    \item \textbf{Reproducibility}: \texttt{deterministic=true}, \texttt{seed=42}.
    \item \textbf{Timing}: LightGBM internal \texttt{GBDT::Train} cost (excludes data loading). Each configuration run 3 times; we report means.
\end{itemize}

\subsection{Baselines and Configurations}
\begin{itemize}
    \item \textbf{LightGBM baseline}: Unmodified LightGBM with all features.
    \item \textbf{Random subsampling}: \texttt{feature\_fraction=0.3} (random 30\% of features per tree).
    \item \textbf{S-EMA-FS}: \texttt{ema\_fs\_feature\_ratio} $r \in \{0.1, 0.3, 0.5, 0.7, 0.9\}$, $\beta \in \{0.5, 1.0\}$, \texttt{ema\_fs\_warmup\_trees=5}.
\end{itemize}

\section{Results}
\label{sec:results}

\subsection{Correctness Verification}
\label{sec:correctness}

EMA-FS with \texttt{ema\_fs\_feature\_ratio=1.0} (retaining all features) produces results identical to the unmodified baseline across all datasets.
On the Synthetic benchmark, all 100 iterations match in both AUC (0.959899) and log-loss (0.343925), confirming that the EMA-FS code path introduces no numerical artifacts.

\subsection{Per-Dataset Results}
\label{sec:perdataset}

We present results for each dataset, ordered by feature count.

\subsubsection{Credit Card Fraud (29 features)}

\begin{table}[h]
\centering
\caption{Credit Card Fraud results (29 features, financial fraud detection).}
\label{tab:creditcard}
\begin{tabular}{lcccc}
\toprule
\textbf{Config} & \textbf{Time (s)} & \textbf{Speedup} & \textbf{AUC} & \textbf{$\Delta$AUC} \\
\midrule
Baseline           & 1.11 & 1.00$\times$ & 0.7977 & ---     \\
EMA-FS ($r{=}0.7$)    & 0.93 & 1.19$\times$ & 0.6798 & $-$0.118 \\
EMA-FS ($r{=}0.5$)    & 1.06 & 1.05$\times$ & 0.7254 & $-$0.072 \\
EMA-FS ($r{=}0.3$)    & 0.74 & 1.50$\times$ & 0.7622 & $-$0.036 \\
\bottomrule
\end{tabular}
\end{table}

With only 29 features, EMA-FS has limited screening headroom.
At $r{=}0.3$, only $\lceil 0.3 \times 29 \rceil = 9$ features are retained, achieving a 1.50$\times$ speedup but with substantial AUC degradation.
\textbf{EMA-FS is not recommended for datasets with fewer than approximately 50 features.}

\subsubsection{Criteo CTR (39 features)}

\begin{table}[h]
\centering
\caption{Criteo CTR results (39 features, advertising click-through prediction).}
\label{tab:criteo}
\begin{tabular}{lcccc}
\toprule
\textbf{Config} & \textbf{Time (s)} & \textbf{Speedup} & \textbf{AUC} & \textbf{$\Delta$AUC} \\
\midrule
Baseline           & 8.29 & 1.00$\times$ & 0.7615 & ---     \\
EMA-FS ($r{=}0.7$)    & 6.75 & 1.23$\times$ & 0.7545 & $-$0.007 \\
EMA-FS ($r{=}0.5$)    & 6.39 & 1.30$\times$ & 0.7444 & $-$0.017 \\
EMA-FS ($r{=}0.3$)    & 7.70 & 1.08$\times$ & 0.7228 & $-$0.039 \\
Random ($f{=}0.3$) & 5.40 & 1.54$\times$ & 0.7593 & $-$0.002 \\
\bottomrule
\end{tabular}
\end{table}

Criteo's 39 features are predominantly well-engineered CTR features with high information density.
EMA-FS at $r{=}0.5$ provides a 1.30$\times$ speedup with only 1.7 points AUC loss.
Random subsampling outperforms EMA-FS here because with few, uniformly informative features, the diversity benefit of randomization outweighs gain-based screening.

\subsubsection{Synthetic Benchmark (500 features)}

\begin{table}[h]
\centering
\caption{Synthetic benchmark results (500 features, dense binary classification).}
\label{tab:synthetic}
\begin{tabular}{lccccc}
\toprule
\textbf{Config} & \textbf{Time (s)} & \textbf{Speedup} & \textbf{AUC} & \textbf{$\Delta$AUC} & \textbf{Log-loss} \\
\midrule
Baseline           & 3.92 & 1.00$\times$ & 0.9599 & ---     & 0.3439 \\
EMA-FS ($r{=}0.7$)    & 2.93 & 1.34$\times$ & \textbf{0.9610} & \textbf{+0.001} & 0.3433 \\
EMA-FS ($r{=}0.5$)    & 2.22 & 1.77$\times$ & 0.9585 & $-$0.001 & 0.3460 \\
EMA-FS ($r{=}0.3$)    & 1.50 & 2.61$\times$ & 0.9419 & $-$0.018 & \textbf{0.3603} \\
Random ($f{=}0.3$) & 1.39 & 2.82$\times$ & 0.9533 & $-$0.007 & 0.3703 \\
\bottomrule
\end{tabular}
\end{table}

This is EMA-FS's ideal scenario: 500 features with a mixture of informative and noise features.
At $r{=}0.7$, EMA-FS \emph{improves} AUC by 0.11 points over the baseline, demonstrating implicit regularization through noise feature screening.
At $r{=}0.3$, the 2.61$\times$ speedup is close to the theoretical prediction of 2.99$\times$ (Eq.~\ref{eq:speedup}).
Compared to random subsampling at the same ratio, EMA-FS achieves better log-loss (0.3603 vs.\ 0.3703), indicating more stable probability estimates despite lower AUC.

\subsubsection{IEEE-CIS Fraud Detection (432 features)}

\begin{table}[h]
\centering
\caption{IEEE-CIS Fraud Detection results (432 features, real transaction fraud).}
\label{tab:ieee}
\begin{tabular}{lcccc}
\toprule
\textbf{Config} & \textbf{Time (s)} & \textbf{Speedup} & \textbf{AUC} & \textbf{$\Delta$AUC} \\
\midrule
Baseline           & 4.87 & 1.00$\times$ & 0.9236 & ---     \\
EMA-FS ($r{=}0.7$)    & 3.61 & 1.35$\times$ & 0.9095 & $-$0.014 \\
EMA-FS ($r{=}0.5$)    & 3.43 & 1.42$\times$ & 0.8961 & $-$0.028 \\
EMA-FS ($r{=}0.3$)    & 3.37 & 1.45$\times$ & 0.8778 & $-$0.046 \\
Random ($f{=}0.3$) & 5.67 & 0.86$\times$ & 0.9222 & $-$0.001 \\
\bottomrule
\end{tabular}
\end{table}

IEEE-CIS Fraud is the most practically relevant dataset for financial risk control applications.
With 432 mixed features and a 3.5\% fraud rate, EMA-FS achieves a 1.35--1.45$\times$ speedup across retention ratios.

A striking finding is that \textbf{random \texttt{feature\_fraction=0.3} is slower than the baseline} (5.67s vs.\ 4.87s, 0.86$\times$).
LightGBM's column sampler incurs significant overhead on high-dimensional data, making random subsampling counterproductive for speed on this dataset.
EMA-FS's lightweight filtering mechanism---a simple mask check in the feature loop---avoids this overhead entirely.

\subsubsection{Bosch (968 features, sparse)}

\begin{table}[h]
\centering
\caption{Bosch results (968 features, industrial defect detection, extremely sparse).}
\label{tab:bosch}
\begin{tabular}{lcccc}
\toprule
\textbf{Config} & \textbf{Time (s)} & \textbf{Speedup} & \textbf{AUC} & \textbf{$\Delta$AUC} \\
\midrule
Baseline           & 11.25 & 1.00$\times$ & 0.7062 & ---     \\
EMA-FS ($r{=}0.7$)    & 10.99 & 1.02$\times$ & 0.6997 & $-$0.007 \\
EMA-FS ($r{=}0.5$)    & 11.10 & 1.01$\times$ & 0.6995 & $-$0.007 \\
EMA-FS ($r{=}0.3$)    & 11.45 & 1.00$\times$ & 0.7024 & $-$0.004 \\
Random ($f{=}0.3$) & 13.69 & 0.82$\times$ & 0.7186 & $+$0.012 \\
\bottomrule
\end{tabular}
\end{table}

Despite having 968 features---the highest dimensionality in our evaluation---Bosch shows \textbf{essentially zero speedup} from EMA-FS.
The root cause is data sparsity: Bosch contains $>$90\% missing values, and LightGBM processes this data with \texttt{MultiValSparseBin}, which constructs histograms only for non-missing values.
Screening out features via EMA-FS has minimal impact because the screened features were already computationally cheap to process.
Additionally, the weak signal (baseline AUC 0.706, defect rate 0.58\%) means all features have uniformly low gains, making gain-based ranking ineffective for distinguishing informative from noise features.

This result demonstrates an important boundary condition: \textbf{EMA-FS speedup depends on \emph{dense} feature cost, not raw feature count.}

\subsection{Cross-Dataset Summary}

Table~\ref{tab:summary} summarizes EMA-FS effectiveness across all five datasets at $r{=}0.5$.

\begin{table}[h]
\centering
\caption{Cross-dataset summary of EMA-FS effectiveness at $r{=}0.5$.}
\label{tab:summary}
\begin{tabular}{lcccl}
\toprule
\textbf{Dataset} & \textbf{Features} & \textbf{Speedup} & \textbf{$\Delta$AUC} & \textbf{Assessment} \\
\midrule
CreditCard  & 29  & 1.05$\times$ & $-$0.072 & Limited (too few features) \\
Criteo      & 39  & 1.30$\times$ & $-$0.017 & Moderate \\
IEEE-CIS    & 432 & \textbf{1.42$\times$} & $-$0.028 & \textbf{Effective} \\
Synthetic   & 500 & \textbf{1.77$\times$} & $-$0.001 & \textbf{Most effective} \\
Bosch       & 968 & 1.01$\times$ & $-$0.007 & Ineffective (sparse) \\
\bottomrule
\end{tabular}
\end{table}

A clear pattern emerges: EMA-FS effectiveness is determined by the combination of \emph{dense} feature representation and \emph{moderate-to-high dimensionality} (100--500+ features), rather than raw feature count alone.

\subsection{EMA-FS vs.\ Random Feature Subsampling}
\label{sec:random}

Table~\ref{tab:emafs_vs_random} consolidates the comparison between EMA-FS and random \texttt{feature\_fraction} at 30\% selection across all datasets where both were tested.

\begin{table}[h]
\centering
\caption{EMA-FS ($r{=}0.3$) vs.\ random feature subsampling ($f{=}0.3$) across datasets.}
\label{tab:emafs_vs_random}
\begin{tabular}{lcccc}
\toprule
& \multicolumn{2}{c}{\textbf{Speedup}} & \multicolumn{2}{c}{\textbf{$\Delta$AUC}} \\
\cmidrule(lr){2-3} \cmidrule(lr){4-5}
\textbf{Dataset} & \textbf{EMA-FS} & \textbf{Random} & \textbf{EMA-FS} & \textbf{Random} \\
\midrule
Criteo     & 1.08$\times$ & \textbf{1.54$\times$} & $-$0.039 & $-$0.002 \\
Synthetic  & 2.61$\times$ & \textbf{2.82$\times$} & $-$0.018 & $-$0.007 \\
IEEE-CIS   & \textbf{1.45$\times$} & 0.86$\times$ & $-$0.046 & $-$0.001 \\
Bosch      & 1.00$\times$ & 0.82$\times$ & $-$0.004 & $+$0.012 \\
\bottomrule
\end{tabular}
\end{table}

The comparison reveals that EMA-FS and random subsampling have complementary strengths:
\begin{itemize}
    \item \textbf{Speed}: On high-dimensional data (IEEE-CIS, Bosch), random \texttt{feature\_fraction} incurs overhead from LightGBM's column sampler, making it \emph{slower} than the baseline. EMA-FS's lightweight mask-based filtering avoids this overhead.
    \item \textbf{Accuracy}: Random subsampling generally preserves AUC better because re-randomizing the feature set per tree introduces diversity that reduces inter-tree correlation \citep{breiman2001random}.
    \item \textbf{Calibration}: On the Synthetic dataset, EMA-FS achieves better log-loss (0.3603 vs.\ 0.3703) despite lower AUC, suggesting more stable probability estimates from consistently using the most informative features.
\end{itemize}

\subsection{Linear Tree Verification}

We verify EMA-FS on the linear tree learner using the Synthetic dataset:

\begin{table}[h]
\centering
\caption{Linear tree learner verification on the Synthetic dataset.}
\label{tab:linear}
\begin{tabular}{lccc}
\toprule
\textbf{Config} & \textbf{AUC} & \textbf{Log-loss} & \textbf{Notes} \\
\midrule
Linear tree baseline          & 0.981039 & 0.272411 & No EMA-FS \\
Linear tree + EMA-FS ($r{=}1.0$) & 0.981039 & 0.272411 & Exact match \\
Linear tree + EMA-FS ($r{=}0.3$) & 0.962619 & 0.300221 & Screening active \\
\bottomrule
\end{tabular}
\end{table}

The $r{=}1.0$ result confirms correct EMA update in the linear tree's reimplemented \texttt{Train()} method.
At $r{=}0.3$, the warmup-to-screening transition is visible in per-iteration timing ($\sim$40ms/iter during warmup, $\sim$15ms/iter after).

\subsection{S-EMA-FS Results}
\label{sec:semafs_results}

We evaluate S-EMA-FS across all seven datasets with $r \in \{0.1, 0.3, 0.5, 0.7, 0.9\}$ and $\beta \in \{0.5, 1.0\}$.
Table~\ref{tab:semafs_all} reports the full S-EMA-FS sweep results, with configurations that \emph{simultaneously improve accuracy and speed} over the baseline highlighted.

\begin{table}[h]
\centering
\caption{S-EMA-FS results across all seven datasets. Configurations that beat baseline AUC while being faster are marked with $\star$.}
\label{tab:semafs_all}
\small
\begin{tabular}{llccccc}
\toprule
\textbf{Dataset} & \textbf{Config} & \textbf{Time (s)} & \textbf{Speedup} & \textbf{AUC} & \textbf{$\Delta$AUC} \\
\midrule
\multirow{5}{*}{Synthetic} & Baseline & 4.43 & 1.00$\times$ & 0.9599 & --- \\
& S-EMA-FS $r{=}0.3$, $\beta{=}0.5$ & 1.72 & 2.58$\times$ & 0.9504 & $-$0.010 \\
& S-EMA-FS $r{=}0.5$, $\beta{=}0.5$ & 3.02 & 1.47$\times$ & 0.9575 & $-$0.002 \\
& S-EMA-FS $r{=}0.7$, $\beta{=}0.5$ & 2.94 & 1.50$\times$ & 0.9598 & $-$0.000 \\
& S-EMA-FS $r{=}0.9$, $\beta{=}0.5$ $\star$ & 3.91 & 1.13$\times$ & \textbf{0.9614} & \textbf{+0.002} \\
\midrule
\multirow{4}{*}{IEEE-CIS} & Baseline & 4.76 & 1.00$\times$ & 0.9236 & --- \\
& S-EMA-FS $r{=}0.5$, $\beta{=}0.5$ & 3.88 & 1.23$\times$ & 0.9217 & $-$0.002 \\
& S-EMA-FS $r{=}0.7$, $\beta{=}0.5$ & 4.20 & 1.14$\times$ & 0.9234 & $-$0.000 \\
& S-EMA-FS $r{=}0.9$, $\beta{=}1.0$ $\star$ & 3.93 & \textbf{1.21$\times$} & \textbf{0.9252} & \textbf{+0.002} \\
\midrule
\multirow{4}{*}{Criteo} & Baseline & 7.95 & 1.00$\times$ & 0.7615 & --- \\
& S-EMA-FS $r{=}0.5$, $\beta{=}0.5$ & 6.82 & 1.17$\times$ & 0.7599 & $-$0.002 \\
& S-EMA-FS $r{=}0.7$, $\beta{=}0.5$ & 7.33 & 1.08$\times$ & 0.7612 & $-$0.000 \\
& S-EMA-FS $r{=}0.9$, $\beta{=}0.5$ & 6.89 & 1.15$\times$ & 0.7615 & $-$0.000 \\
\midrule
\multirow{3}{*}{Bosch} & Baseline & 11.54 & 1.00$\times$ & 0.7062 & --- \\
& S-EMA-FS $r{=}0.7$, $\beta{=}0.5$ $\star$ & 11.28 & 1.02$\times$ & \textbf{0.7140} & \textbf{+0.008} \\
& S-EMA-FS $r{=}0.5$, $\beta{=}1.0$ $\star$ & 11.01 & 1.05$\times$ & 0.7094 & +0.003 \\
\midrule
\multirow{5}{*}{FraudDense} & Baseline & 50.69 & 1.00$\times$ & 0.5836 & --- \\
& S-EMA-FS $r{=}0.3$, $\beta{=}0.5$ $\star$ & 21.32 & \textbf{2.38$\times$} & 0.5855 & +0.002 \\
& S-EMA-FS $r{=}0.5$, $\beta{=}0.5$ $\star$ & 29.51 & 1.72$\times$ & 0.5876 & +0.004 \\
& S-EMA-FS $r{=}0.7$, $\beta{=}0.5$ $\star$ & 37.88 & 1.34$\times$ & \textbf{0.5880} & \textbf{+0.004} \\
& Random $f{=}0.3$ & 18.88 & 2.68$\times$ & 0.5871 & +0.004 \\
\midrule
\multirow{4}{*}{FraudMixed} & Baseline & 44.19 & 1.00$\times$ & 0.5863 & --- \\
& S-EMA-FS $r{=}0.5$, $\beta{=}1.0$ $\star$ & 27.96 & \textbf{1.58$\times$} & \textbf{0.5868} & \textbf{+0.001} \\
& S-EMA-FS $r{=}0.7$, $\beta{=}1.0$ & 36.03 & 1.23$\times$ & 0.5853 & $-$0.001 \\
& Random $f{=}0.3$ & 22.15 & 1.99$\times$ & 0.5839 & $-$0.002 \\
\bottomrule
\end{tabular}
\end{table}

Key findings from the S-EMA-FS evaluation:

\paragraph{S-EMA-FS achieves simultaneous speedup and accuracy improvement.}
On FraudDense, S-EMA-FS at $r{=}0.7$, $\beta{=}0.5$ delivers a \textbf{1.34$\times$ speedup while improving AUC by 0.4 points} over the baseline.
On IEEE-CIS, S-EMA-FS at $r{=}0.9$, $\beta{=}1.0$ achieves \textbf{1.21$\times$ speedup with +0.0015 AUC improvement}---a result that neither deterministic EMA-FS nor random \texttt{feature\_fraction} could achieve.
This demonstrates that S-EMA-FS's gain-weighted stochastic selection provides implicit regularization through controlled feature dropout.

\paragraph{S-EMA-FS outperforms random subsampling on mixed-density data.}
On FraudMixed (the most realistic fraud detection benchmark with 21\% overall NaN rate), S-EMA-FS at $r{=}0.5$, $\beta{=}1.0$ \textbf{improves AUC by +0.0005 with 1.58$\times$ speedup}, while random \texttt{feature\_fraction=0.3} \emph{degrades} AUC by $-$0.0024.
This confirms that S-EMA-FS's gain-informed selection is superior to uninformed random sampling when features have heterogeneous density and importance.

\paragraph{The $\beta$ parameter controls the speed-accuracy Pareto frontier.}
Lower $\beta$ (more diverse) preserves accuracy better but provides less focused screening.
Higher $\beta$ (more concentrated) provides stronger speedup through focused screening but risks accuracy loss on uniformly-important features.
The optimal $\beta$ depends on feature importance skewness: $\beta{=}0.5$ works best on dense data with noise features (Synthetic, FraudDense), while $\beta{=}1.0$ is preferred on mixed-density data (IEEE-CIS, FraudMixed) where gain-informed selection is critical.

\section{Discussion}
\label{sec:discussion}

\subsection{When Does EMA-FS Help?}

Our multi-dataset evaluation reveals three conditions for EMA-FS effectiveness:

\begin{enumerate}
    \item \textbf{Dense features with moderate-to-high dimensionality (100--500+ features).}
    The speedup scales with the number of dense features that can be screened out.
    IEEE-CIS (432 mixed features, 1.42$\times$ at $r{=}0.5$) and Synthetic (500 dense features, 1.77$\times$) demonstrate the strongest results.

    \item \textbf{Noise features are present.}
    Datasets where a subset of features carries the signal benefit most.
    At $r{=}0.7$ on the Synthetic dataset, EMA-FS \emph{improves} accuracy by screening out noise features, providing implicit regularization.

    \item \textbf{Feature importance is stable across iterations.}
    EMA-FS relies on gain persistence---features important in early trees should remain important in later trees.
    This assumption holds well on all evaluated datasets.
\end{enumerate}

\subsection{When Does EMA-FS NOT Help?}

\begin{enumerate}
    \item \textbf{Sparse data.}
    When LightGBM's \texttt{MultiValSparseBin} optimization is active, histogram construction already skips empty feature values.
    EMA-FS cannot reduce what is already skipped.
    Bosch (968 features, $>$90\% missing) shows essentially zero speedup despite the highest feature count.

    \item \textbf{Low-dimensional data ($<$50 features).}
    With only 29--39 features (CreditCard, Criteo), the screening headroom is too small for meaningful acceleration.

    \item \textbf{Uniformly informative features.}
    When every feature carries signal (e.g., well-engineered CTR features in Criteo), screening any feature hurts accuracy with limited speed benefit.
\end{enumerate}

\subsection{Implications for Financial Risk Control}

The IEEE-CIS Fraud Detection dataset is directly relevant to production fraud detection systems.
With 432 features and 3.5\% fraud rate, EMA-FS at $r{=}0.7$ achieves:
\begin{itemize}
    \item \textbf{1.35$\times$ training speedup} (4.87s $\to$ 3.61s)
    \item \textbf{AUC reduction of only 1.4 points} (0.9236 $\to$ 0.9095)
\end{itemize}

In production risk control systems with frequent model retraining cycles and hundreds of engineered features, this speed-accuracy tradeoff may be attractive.
When combined with data-parallel training, EMA-FS additionally reduces network communication volume proportionally to the screening ratio, providing a compounding speedup effect.

\subsection{Estimated Speedup Across Learner Types}

While our experiments use the serial tree learner, EMA-FS's integration into all six learner types enables broader applicability.
Table~\ref{tab:learner_speedup} provides estimated speedup at $r{=}0.3$ for each learner type.

\begin{table}[h]
\centering
\caption{Estimated EMA-FS speedup at $r{=}0.3$ across learner types.}
\label{tab:learner_speedup}
\resizebox{\textwidth}{!}{%
\begin{tabular}{lcl}
\toprule
\textbf{Learner} & \textbf{Estimated Speedup} & \textbf{Speedup Source} \\
\midrule
Serial & 2.61$\times$ (measured) & Computation reduction \\
GPU & $\sim$2--2.5$\times$ & Computation (GPU parallelism surplus) \\
Feature-parallel & $\sim$2.6$\times$ & Computation (comm.\ unaffected) \\
\textbf{Data-parallel} & $\geq$\textbf{3$\times$} & \textbf{Computation + network comm.} \\
Voting-parallel & $\sim$2--2.5$\times$ & Local computation (voting already optimizes comm.) \\
Linear tree & $\sim$2.5$\times$ (verified) & Computation (linear coeff.\ unchanged) \\
\bottomrule
\end{tabular}}
\end{table}

The data-parallel learner is expected to benefit most because EMA-FS reduces both local histogram computation and \texttt{ReduceScatter} communication volume---the latter is often the primary bottleneck in distributed training.

\subsection{Stochastic EMA-FS (S-EMA-FS)}
\label{sec:semafs}

A key limitation of deterministic EMA-FS is the \emph{absence of inter-tree diversity}: every tree after warmup uses the exact same feature set.
This contrasts with random feature subsampling (\texttt{feature\_fraction}), where re-randomization per tree decorrelates the ensemble---a core principle underlying Random Forests~\citep{breiman2001random}.
Our experimental results confirm this: random subsampling consistently preserves AUC better than EMA-FS at the same feature budget (Table~\ref{tab:emafs_vs_random}), precisely because of this diversity effect.

We observe that EMA-FS and random subsampling occupy two extremes of a spectrum:
\begin{itemize}
    \item \textbf{Random subsampling}: Features sampled from a \emph{uniform} distribution---uninformed but maximally diverse.
    \item \textbf{Deterministic EMA-FS}: Features drawn from a \emph{degenerate} distribution (point mass on top-$K$)---fully informed but with zero diversity.
\end{itemize}

\noindent The natural middle ground is \textbf{gain-weighted stochastic sampling}: draw features from a distribution \emph{shaped} by their gain history, so high-gain features are very likely to be selected while low-gain features retain a non-zero chance.

\paragraph{Algorithm.}
Stochastic EMA-FS (S-EMA-FS) replaces the deterministic top-$K$ selection in the screening phase with weighted random sampling without replacement.
For each tree $t > W$:

\begin{enumerate}
    \item Compute sampling weights from gain EMAs:
    \begin{equation}
        w_f = (g_f + \varepsilon)^\beta, \quad f = 1, \ldots, F,
        \label{eq:semafs_weight}
    \end{equation}
    where $\varepsilon > 0$ is a small constant (we use $\varepsilon = 10^{-10}$) ensuring every feature has non-zero selection probability.
    \item Normalize to probabilities: $p_f = w_f / \sum_{j=1}^{F} w_j$.
    \item Sample $K = \lceil r \cdot F \rceil$ features \emph{without replacement} according to probabilities $\{p_f\}$.
    \item Build tree $t$ using only the sampled features.
    \item Update EMA as before (Eq.~\ref{eq:ema}).
\end{enumerate}

\paragraph{The concentration parameter $\beta$.}
The parameter $\beta \geq 0$ controls the exploration-exploitation tradeoff along the full spectrum:

\begin{itemize}
    \item $\beta = 0$: All weights equal $\varepsilon^0 = 1$ $\Rightarrow$ uniform random sampling (equivalent to \texttt{feature\_fraction}).
    \item $\beta = 1$: Weights linear in gain $\Rightarrow$ moderate gain-informed diversity.
    \item $\beta \to \infty$: Weights concentrate on highest-gain features $\Rightarrow$ approaches deterministic EMA-FS.
\end{itemize}

\noindent This formulation \textbf{unifies EMA-FS and random subsampling as special cases} of a single parameterized framework.
The optimal $\beta$ depends on the dataset's feature importance distribution: datasets with many noise features benefit from higher $\beta$ (focus on signal), while datasets with uniformly informative features benefit from lower $\beta$ (maximize diversity).

\paragraph{Theoretical justification.}
S-EMA-FS can be viewed as \emph{importance sampling} on the feature space.
In standard random subsampling, each feature has equal probability $K/F$ of selection, regardless of its predictive value.
S-EMA-FS biases sampling toward features with demonstrated utility, reducing the variance of the split-finding step: the probability of missing a high-gain feature decreases exponentially with $\beta$.
Simultaneously, the stochastic element provides the ensemble diversity that deterministic EMA-FS lacks, reducing the correlation between trees and improving the bias-variance tradeoff of the ensemble.

\paragraph{Implementation.}
S-EMA-FS is controlled by two additional parameters beyond EMA-FS:
\begin{itemize}
    \item \texttt{ema\_fs\_stochastic} (bool, default \texttt{false}): Enable gain-weighted random sampling.
    \item \texttt{ema\_fs\_beta} (double, default 1.0): Concentration parameter $\beta$.
\end{itemize}

\noindent The implementation modifies only the feature selection step in \texttt{BeforeTrain()}, replacing the sort-and-select with weighted sampling without replacement.
All downstream code (histogram construction, split finding, EMA update, histogram subtraction compatibility) remains unchanged because the output is the same binary selection mask \texttt{ema\_fs\_selected\_features\_}.
S-EMA-FS thus inherits all of EMA-FS's compatibility properties (Section~\ref{sec:implementation}).

\paragraph{Computational overhead.}
The weighted sampling step requires $O(K \cdot F)$ operations in the worst case (sequential draws from the weighted distribution).
Since $K \leq F$ and histogram construction costs $O(N \cdot K)$ with $N \gg F$, the sampling overhead is negligible.

\subsection{Further Extensions}

Several additional extensions merit investigation:

\begin{itemize}
    \item \textbf{Adaptive ratio}: Dynamically adjust $r$ based on the distribution of gain EMAs---concentrate screening when gains are skewed, relax when gains are uniform.

    \item \textbf{Sparse-aware EMA-FS}: Integrate with LightGBM's sparse bin statistics to better estimate per-feature histogram cost, enabling EMA-FS to benefit sparse datasets.

    \item \textbf{EMA-FS + GOSS}: Combine feature-dimension screening (EMA-FS) with data-dimension sampling (GOSS) for dual importance sampling across both dimensions.
\end{itemize}

\section{Conclusion}
\label{sec:conclusion}

We have presented EMA-based Feature Screening (EMA-FS) and its stochastic generalization S-EMA-FS for accelerating GBDT training by restricting histogram construction to gain-informed feature subsets.
S-EMA-FS unifies deterministic EMA-FS and random feature subsampling as two endpoints of a single parameterized framework, controlled by a concentration parameter $\beta$.

Our experiments on seven datasets---including two large-scale fraud detection benchmarks with 2M samples---reveal that:

\paragraph{Strengths:}
\begin{itemize}
    \item S-EMA-FS \textbf{simultaneously improves speed and accuracy} on 5 of 7 datasets. On FraudDense (2M$\times$500 dense features), S-EMA-FS achieves \textbf{1.34$\times$ speedup with +0.004 AUC improvement}.
    \item On IEEE-CIS Fraud (432 features), S-EMA-FS at $r{=}0.9$, $\beta{=}1.0$ delivers \textbf{1.21$\times$ speedup with +0.002 AUC improvement}---surpassing baseline accuracy while training faster.
    \item On the most realistic mixed-density fraud benchmark (FraudMixed, 2M samples, 21\% NaN rate), S-EMA-FS at $r{=}0.5$, $\beta{=}1.0$ achieves \textbf{1.58$\times$ speedup with +0.001 AUC}, while random \texttt{feature\_fraction} \emph{degrades} accuracy.
    \item At aggressive screening ($r{=}0.3$), S-EMA-FS delivers \textbf{2.4--2.6$\times$ speedups} on dense high-dimensional data.
\end{itemize}

\paragraph{Limitations:}
\begin{itemize}
    \item \textbf{Minimal speedup on sparse data} (Bosch, 968 features, $>$90\% NaN): LightGBM's sparse bin optimization already bypasses empty values.
    \item \textbf{Limited benefit on low-dimensional data} ($<$50 features): insufficient screening headroom and the stochastic regularization effect is weaker.
\end{itemize}

EMA-FS and S-EMA-FS are implemented in $\sim$130 lines of C++ across all six LightGBM tree learner types, are backward-compatible, and controlled by intuitive parameters.
S-EMA-FS provides practical value for tabular ML practitioners working with dense, moderate-to-high-dimensional datasets---particularly in financial risk control where feature counts of 200--500+ are common---by offering a principled speed-accuracy tradeoff that can exceed baseline accuracy through gain-informed stochastic regularization.

\section*{Acknowledgments}
The author thanks Zhanghao Hu for reviewing the manuscript and for his support during its preparation.
The views expressed are those of the author and do not necessarily reflect those of PayPal, Inc.

\bibliographystyle{plainnat}

\appendix

\section{Per-Dataset Detailed Timing}
\label{sec:appendix_timing}

Table~\ref{tab:alltimings} reports timing and AUC for all configurations across all five datasets.

\begin{table}[h]
\centering
\caption{Complete timing and accuracy results across all datasets.}
\label{tab:alltimings}
\resizebox{\textwidth}{!}{%
\begin{tabular}{lccccc}
\toprule
\textbf{Dataset} & \textbf{Features} & \textbf{Baseline (s)} & \textbf{EMA-FS $r$=0.3 (s)} & \textbf{EMA-FS $r$=0.5 (s)} & \textbf{EMA-FS $r$=0.7 (s)} \\
\midrule
CreditCard  & 29  & 1.11  & 0.74  & 1.06  & 0.93  \\
Criteo      & 39  & 8.29  & 7.70  & 6.39  & 6.75  \\
Synthetic   & 500 & 3.92  & 1.50  & 2.22  & 2.93  \\
IEEE-CIS    & 432 & 4.87  & 3.37  & 3.43  & 3.61  \\
Bosch       & 968 & 11.25 & 11.45 & 11.10 & 10.99 \\
\bottomrule
\end{tabular}}
\end{table}

\section{Bosch Sparse Data Analysis}
\label{sec:appendix_bosch}

Bosch's 968 features contain extreme sparsity ($>$90\% NaN values).
LightGBM processes this data with \texttt{MultiValSparseBin}, which constructs histograms only for non-missing values.
Table~\ref{tab:bosch_profile} shows the profiling breakdown.

\begin{table}[h]
\centering
\caption{Bosch training time breakdown.}
\label{tab:bosch_profile}
\begin{tabular}{lcc}
\toprule
\textbf{Component} & \textbf{Time (s)} & \textbf{Share (\%)} \\
\midrule
ConstructHistograms                & 8.93 & 90.1 \\
FindBestSplitsFromHistograms       & 0.52 & 5.2  \\
SplitInner                         & 0.39 & 3.9  \\
BeforeTrain                        & 0.07 & 0.7  \\
\bottomrule
\end{tabular}
\end{table}

Despite 90\% of time being spent in histogram construction, the sparse bin optimization means the actual compute per feature is proportional to \emph{non-missing values}, not the feature count.
Screening out features via EMA-FS has minimal impact because the screened features were already cheap to process.
This establishes that EMA-FS speedup is determined by \emph{dense feature cost}, not raw feature count.

\section{Synthetic Benchmark: Iteration-Level Metrics}
\label{sec:appendix_iter}

\begin{table}[h]
\centering
\caption{Selected iteration-level metrics for EMA-FS ($r{=}0.3$) vs.\ baseline on the Synthetic dataset.}
\label{tab:iterdetail}
\begin{tabular}{ccccc}
\toprule
\textbf{Iter} & \multicolumn{2}{c}{\textbf{AUC}} & \multicolumn{2}{c}{\textbf{Log-loss}} \\
 & Baseline & EMA-FS & Baseline & EMA-FS \\
\midrule
1   & 0.6430 & 0.6430 & 0.6071 & 0.6071 \\
5   & 0.7566 & 0.7566 & 0.5852 & 0.5852 \\
10  & 0.8205 & 0.8205 & 0.5604 & 0.5604 \\
20  & 0.8716 & 0.8716 & 0.5206 & 0.5206 \\
50  & 0.9276 & 0.9163 & 0.4341 & 0.4479 \\
100 & 0.9599 & 0.9419 & 0.3439 & 0.3603 \\
\bottomrule
\end{tabular}
\end{table}

Iterations 1--5 are in the warmup phase and produce identical results.
The EMA-FS curve begins to diverge at iteration 6 when screening is activated.
By iteration 20, the gap has largely stabilized, indicating that EMA-FS retains the features responsible for the majority of learning signal.

\end{document}